\documentclass[11pt,letterpaper]{article}
\usepackage{naaclhlt2016}
\usepackage{times}
\usepackage{latexsym}

\usepackage{graphicx}
\usepackage{color}
\usepackage{xcolor}
\usepackage{tikz}
\usepackage{amsmath}
\usepackage{caption}
\usepackage{paralist}
\usepackage{subfig}
\usepackage{booktabs}
\usepackage{amssymb}
\usepackage{amsmath}
\usepackage{enumerate}
\usepackage{xspace}
\usepackage[hyphens]{url}
\usepackage{multirow,paralist}
\usepackage{hyperref}
\usepackage{makeidx}
\naaclfinalcopy 
\def\naaclpaperid{***} 

\newcommand{\figref}[2][]{Figure#1~\ref{#2}\xspace}
\newcommand{\tabref}[2][]{Table#1~\ref{#2}\xspace}

\newcommand{\WE}[1][]{\textsc{WEmb}#1\xspace}
\newcommand{\MWE}[1][]{\textsc{MWE}#1\xspace}

\newcommand{\relation}[2][]{\texttt{#2}$_{\texttt{#1}}$\xspace}
\newcommand{\ex}[1]{\textit{#1}\xspace}
\newcommand{\sense}[1]{\textsc{#1}\xspace}
\newcommand{\gap}{\relation{Gap}}
\newcommand{\pardist}{\relation[Dist]{Par}}
\newcommand{\parparent}{\relation[Parent]{Par}}
\newcommand{\interqt}{\relation{InterQt}}
\newcommand{\fstcap}{\relation[First]{Cap}}
\newcommand{\capnorm}{\relation[Norm]{Cap}}
\newcommand{\capratio}{\relation[Ratio]{Cap}}
\newcommand{\hasnonalpha}{\relation[NAlpha]{Has}}
\newcommand{\hasnum}{\relation[Num]{Has}}
\newcommand{\hasprime}{\relation[Prime]{Has}}
\newcommand{\isat}{\relation[At]{Is}}
\newcommand{\ishash}{\relation[Hash]{Is}}
\newcommand{\isnum}{\relation[Num]{Is}}
\newcommand{\ispunct}{\relation[Punct]{Is}}
\newcommand{\isunknown}{\relation[Unk]{Is}}
\newcommand{\isurl}{\relation[Url]{Is}}
\newcommand{\prequoted}{\relation[Pre]{Quot}}
\newcommand{\postquoted}{\relation[Post]{Quot}}
\newcommand{\logrange}{\relation[Range]{Log}}
\begin{document}
\title{VectorWeavers at
SemEval-2016 Task 10: From Incremental Meaning to Semantic Unit\\ (phrase by phrase)
}
\author{Andreas Scherbakov \qquad Ekaterina Vylomova \qquad Fei Liu \qquad Timothy Baldwin \\
	    {\tt andreas@softwareengineer.pro, evylomova@gmail.com,}\\{\tt
  fliu3@student.unimelb.edu.au, tb@ldwin.net} \\[1ex]
  The University of Melbourne\\
  VIC 3010, Australia}

\date{}

\maketitle

\begin{abstract}
This paper describes an experimental approach to Detection of Minimal Semantic Units and their Meaning (DiMSUM), explored within the framework of SemEval'16 Task 10. The approach is primarily based on a combination of word embeddings and parser-based features, and employs unidirectional incremental computation of compositional embeddings for multiword expressions.
\end{abstract}

\section{Motivation}

This paper proposes an approach to the segmentation of POS-tagged English sentences into minimal semantic units along with labelling of units with semantic classes (``supersenses'').
Supersenses are ``lightweight semantic annotation[s] of text originating in WordNet'' \cite{schneider2012coarse}. Here, we investigate two major ideas, as follows.

First, inspired by \newcite{salehi2015word} we hypothesise that word embeddings (\WE[s]), e.g.\ from word2vec \cite{mikolov2013distributed}, could be an extremely valuable resource of knowledge for guessing the sense of the word. \WE[s] have been shown to represent distinguishable sense components learnt from large training corpora. Many papers have described experiments with word meaning extraction from word embeddings, and demonstrated that it's possible to detect semantic relations between words based on them. Additionally, they may be used to simulate various types of relations between words with simple linear operations over word vectors, and for capturing morphological properties of words \cite{Mikolov+:2013,Vylomova+:2015}. However, more generally, they lack any coherent sense-to-value correspondence.
According to this, it's natural to attempt to use \WE[s] as a source of word meaning information in predicting the senses of semantic units, as part of the larger task of simultaneously detecting minimal semantic units and assigning them meaning. It should be noted that the general term ``multiword expression'' (\MWE) embraces expressions of different types \cite{Baldwin:Kim:2009a}. Some of them are meaning-based and the meaning may be preserved under substitution with synonyms, e.g.; other kinds, on the other hand, are semantically idiomatic word combinations, where we would expect \WE[s] to have less utility. This first approach is thus an examination of the use of \WE[s] in analysing semantic units of mixed semantic compositionality.

Our second hypothesis is that parser-based features will positively impact on the identification of \MWE[s].
Our preliminary explorations showed that the syntactic structure of a text is closely related to the  probability of an \MWE occurring. For instance, in  almost all cases an \MWE is fully subsumed within  a single clause; in the DiMSUM training set, e.g., there is just one sentence (out of 4800) where this condition is violated.  Additionally, all of the components of an \MWE tend to be directly connected within a dependency graph. As we observe strong correlation between the distance between two words in a parse tree and their likelihood of forming an \MWE, we decided to employ parse trees as a source of features. It may be noted that we initially attempted to create a system where an \MWE is treated as a special kind of clause: we modified the syntactic tree using knowledge of \MWE boundaries in the training corpus and then trained a special version of the parser aware of such modified trees. That special parser version was used to directly produce \MWE-labeled clauses over an arbitrary text. Although such a direct approach didn't yield good results, we believe that directly incorporating \MWE identification as part of the parsing process is a promising and fruitful direction for future work.

In our submission, we intentionally avoided the use of any pre-existing lexical resources, including multiword lexicons, to better focus our attention on \WE[s]. 

Our source code is available at:\\ \url{https://github.com/andreas-softwareengineer-pro/dimsum-semeval2016}.

\begin{figure*}[t]
\begin{center}
\includegraphics [scale=0.28]{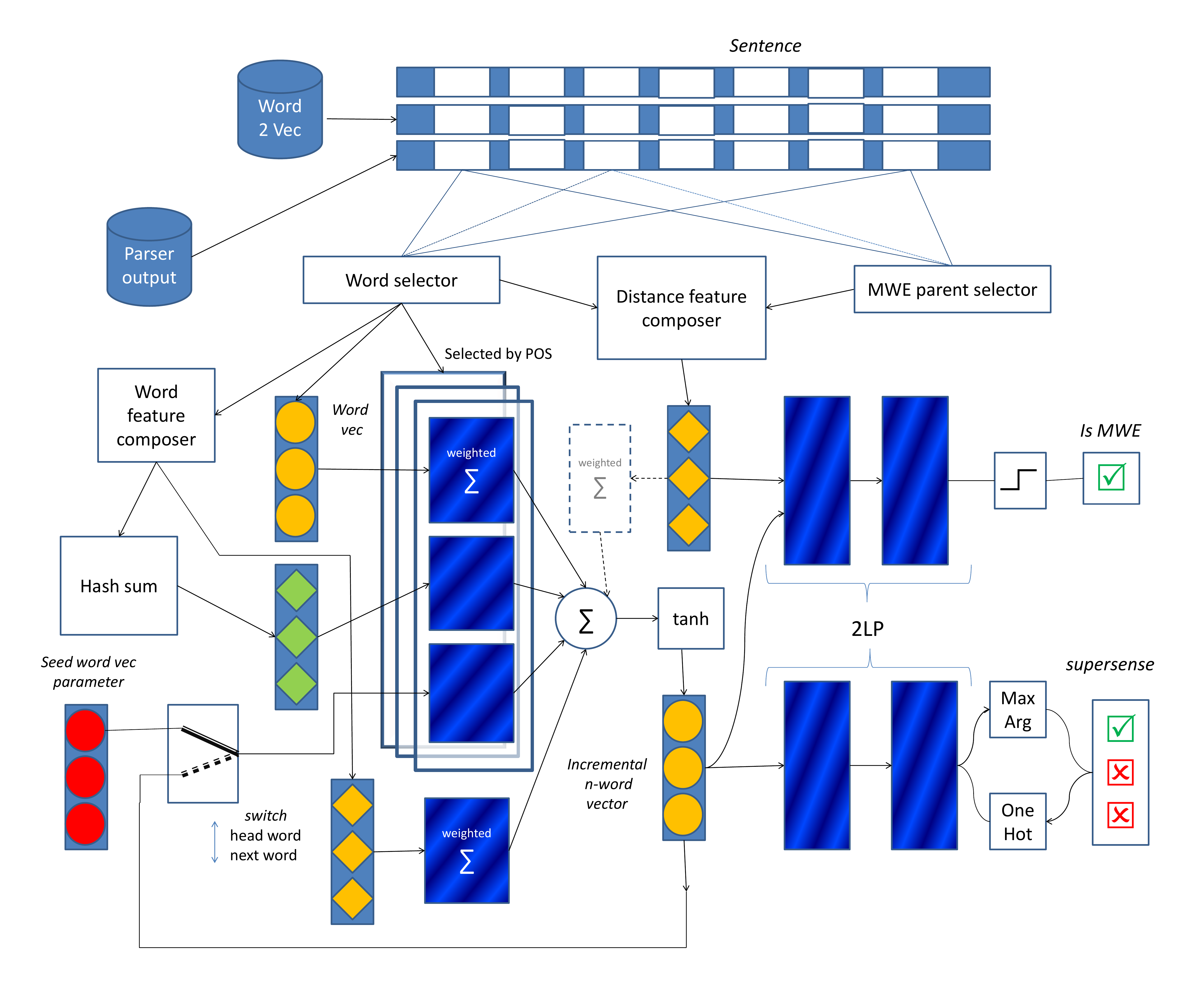}
\end{center}
\caption{An outline of the system architecture}
\label{arch}
\end{figure*}

\section{Overview}

An overall architecture of our system is shown in \figref{arch}. The neural network consists of three blocks: (1) an \textit{incremental vector (recurrency)} calculator for a candidate \MWE; (2) a two-layer perceptron for sense classification; and (3) a (somewhat wider) two-layer perceptron for \MWE classification. An incremental vector produced by the first block is used as a source of feature vectors by each of the latter blocks. 

The  incremental vector calculator produces a vector $v_n\in{\rm I\!R}^{D_v}$ for a $n$-word semantic unit expression:
\begin{equation*}
v_n = \tanh\left(\begin{array}{l}
W_v(P_n) \times wordvec(w_n) + \\
W_h(P_n) \times hash(w_n) + \\
W_f\times wordfeat(w_n) + \\
W_c(P_n)\times\left\lbrace\begin{array}{ll}
v_{n-1}, & n>0 \\
seed, & n=0 \end{array}\right\rbrace\\
\end{array}\right)
\end{equation*}
where $W_v\in {\rm I\!R}^{D_v\times M}$,  $W_h\in {\rm I\!R}^{D_h\times M}$, $W_f\in {\rm I\!R}^{D_f\times M}$ and  $W_c\in {\rm I\!R}^{D_c\times M}$ are parameter matrices, $D_v$, $D_h$, $D_f$ and $D_c$ are their respective feature vector sizes, $M$ is the incremental vector size, $w_n$ and $P_n$ are the $n$th word and its part of speech, respectively, $wordvec$ is a word embedding lookup, $hash$ is a hash function over characters of the word (used as a means of generating embeddings for unknown words and preserve the ability to distinguish concrete words), and $wordfeat$ produces a vector of various word-wise features (see \tabref{wordfeat}), in the spirit of approaches like \newcite{hajivc2009semi}. When calculating $v_1$ (for a single word), an initial $seed$ vector of parameters learnt through backpropagation is used. The motivation behind using the $n=1$ stage for every \MWE (rather than simply starting with $n=2$ and two word vectors) is based on an intention to avoid switching between word embeddings (that play the role of features) and internally calculated $v_i$ that we expect to better capture the structure of the \MWE based on the \WE[s] (including following their dimension counts).  Also, this technique makes the vector composition learning cycle more frequent w.r.t.\ the amount of training data, improving the learning rate and the final penalty.  This evaluation schema is inspired by the work of Socher et al.\ \shortcite{socher2011parsing,socher2010learning,socher2014recursive}. It's actually a recursive neural network (RNN) but, in contrast to previously used techniques, the recursion is based on candidate \MWE[s] rather than the whole content of the sentence.

The incremental vector is used as an input to two two-layer perceptrons: one for \MWE classification and one for one-hot sense vector learning. Back propagation processing of these two also drives the training of the incremental vector calculator, as described above.

We calculate distance-based features based on positions of two adjacent words in an \MWE. A \textit{position} here may mean a position in a sentence, in a parse tree, or, say, inside or outside a quoted phrase. The distance feature vector is supplied as input to the \MWE classifier. As an alternative, it may be supplied to the incremental vector calculator input (shown by the dotted line at \figref{arch}, although no significant difference in performance was observed when we did this).

\section{Learning}
As briefly mentioned above, our procedure considers $n$-word expressions incrementally,  starting with single words.

\subsection{MWE boundary}
The system learns a single scalar output value of +1 for every extension from $(n-1)$-word prefix to a $n$-word \MWE (either complete or incomplete), where $n\geq 2$. In such a way, every $n$-word expression finally yields $n-1$ incremental positive samples.

Also, for every word, regardless of whether it is a standalone word or a member of \MWE,  it learns a $-1$ value (i.e.\ a negative sample) for its (unobserved) extension to an \MWE with a random word. Such a random word is chosen of the $L$ words ahead in the same sentence, not involved in the same \MWE with the current word.\footnote{We limit $L$ to 9 words, and also limit the probability of every word selection to $<0.25$ (by randomly down-weighting the sample processing for $L<4$) in order to prevent possible bias to the distribution of inter-word offsets occurring while processing the last few words of a sentence.  We  solely pick the same sentence words as negative samples in order to make learning  more focused on prediction environment patterns rather than noise, as we use a small training corpus, and at the same time to keep an approximate balance in count between negative and positive samples.}

During the learning process (and, correspondingly, during the prediction), recursive computations of the incremental vector are solely fed by members of the same \MWE, and are restarted for each semantic unit.

\subsection{Senses}
We learn a sense once per complete $n$-word expression, including $n=1$. We have a distinct output per sense, including ``unknown'' (42 in total in the DiMSUM data set). We factorise sense classification error by the number of senses in order to balance the backpropagation of two perceptrons to the incremental vector calculator.

\section{Prediction}
We use a greedy procedure for \MWE prediction in a text. An \textit{outer} loop iterates through all words of a sentence. It selects  each word (not yet consumed by some previously predicted \MWE) as the head word of a possible new \MWE and restarts the incremental vector computation. An \textit{inner} loop iterates through (up to $L$) remaining words in the sentence following the current head word. Each time, a probability for such a second word to be a continuation of the \MWE is evaluated. Once a new \MWE or an \MWE extension is predicted, it consumes the right hand word and designates it to be the next word of the \MWE; the incremental vector is updated respectively with the new \MWE member. Such a procedure is able to generate a deep stack of nested \MWE[s] with gaps, but we restrict the depth to be compliant to the DiMSUM data format. 

\section{Features}
\subsection{Word Embeddings}
We utilised a publicly available pre-trained CBOW word2vec model, with 300-dimensional word vectors based on the Google News corpus. Out-of-vocabulary words are represented with zero-filled vectors.

We use the following search strategy when looking up a word in the word2vec dictionary (until the first successful lookup returns the final embedding):\footnote{We tried using light spelling correction to reduce the rate of unknown words (a simple substitution of one character and then selecting the most probable word according to its frequency  ranking). However it didn't seem to be effective, as the number of mispredicted words was greater then the number of correctly predicted ones, in particular due to the absence of a great number of frequent words (``stop-words'') in the Google News vectors DB. Thus, a more sophisticated system would be needed if one wants to correct typos.}.
\begin{enumerate}\itemsep0pt
\item strip off leading ``\#'' and ``@''
\item if the word is a number then replace by ``NUM''
\item lowercase the word
\item (optionally) lemmatize 
\item remove all non-alpha characters
\item return the word embedding associated with the word OR if no match, return a zero vector
\end{enumerate}

\subsection{Word Hash Sum}
We produce a 64-dimensional $+1$/$-1$ hash sum vector for words where the embedding vector is unknown; a zero vector was supplied for words (somehow) found in the Google News \WE database. Only alphabetic characters were counted in the hash sum.

\subsection{Word Distance features}
\tabref{distfeat} displays inter-word distance-based features.

We used a parser prediction file created for the source text in order to evaluate a hierarchical distance between two candidate words. The hierarchical distance here means the maximum of two counts of edges that connect two given words to the nearest clause they share. For instance, two sibling nodes have the hierarchical distance value of one. The same is also true of a phrase head word and its immediate dependent word (that case is indicated by a distinct feature).
We employed TurboParser v.2.0.0 \cite{martins2014priberam} and trained it on a Penn TreeBank data collection (some conversion was needed to match with the part-of-speech tagset used in the DiMSUM task).

\begin{table}[t]
\begin{tabular}{p{1.45cm}p{0.9cm}p{4cm}}
\hline
Feature&Range&Description\\
\hline
\gap & $\frac{n}{8}$,$n\in Z_{\geq 0} $ & gap between word positions (divided by 8) \\
\pardist & $\lbrace 2;0; $ $-\frac{3}{2} \rbrace $ & hierarchical distance  between words in a parse tree; three gradations for distances of 1,2,$\geq $ 3 \\
\parparent & $\{-1,1\}$ &  two words are head word and child word of the same clause, according to the parser output \\
\interqt& $\{-1,1\}$ & double-quotes anywhere between given words? \\
\hline
\end{tabular}
\caption{Distance features}
\label{distfeat}
\end{table}

\subsection{Heuristic word features}
The full list of miscellaneous word features (concerning capitalization, punctuation characters, lookup success etc.) is presented in \tabref{wordfeat}. 

\begin{table}[t!]
\centering
\begin{tabular}{p{1.4cm}p{0.95cm}p{4cm}}
\hline
Feature   & Value range                                                                                         & Description  \\
\hline
\fstcap & $\{-1,1\}$ & Word starts with a capital letter \\
\capnorm & $[0,1]$ & Word starts with a capital letter, a value normalized by dividing by the sentence word count \\
\capratio & $[0,1]$ &	Ratio of uppercase letters in the word   \\
\hasnonalpha & $\{-1,1\}$&	Word has non-alpha characters  \\
\hasnum 	& $\{-1,1\}$ &Word has any digit inside \\
\hasprime & $\{-1,1\}$ & Word has an apostrophe \\
\isat	& $\{-1,1\}$ & Word starts with ``@''  \\
\ishash	& $\{-1,1\}$ &Word starts with ``\#''  \\
\isnum	& $\{-1,1\}$ & Word is num (integer, float or ``NUM'' keyword)    \\
\ispunct & $\{-1,1\}$ & Word contains punctuation character(s) (one or more of ``!?.,;:{}[]()/'' ) \\
\isunknown	& $\{-1,1\}$ & word2vec out-of-vocabulary word, including stop words  \\
\isurl  & $\{-1,1\}$ &	Word is ``URL'' \\
\logrange	& $[0,12]$ & Smoothed logarithm of the word's frequency range found in the word2vec dictionary, $\log(0.1 \times \text{range} + 1)$; the more frequent the word, the lower the value \\
\prequoted	& $\{-1,1\}$ & A double-quote is located at the previous word position \\
\postquoted & $\{-1,1\}$ & A double-quote is  found at the next word position \\
\hline
\end{tabular}
\caption{Word Features}
\label{wordfeat}
\end{table}

\begin{table}[t!]
\centering
\begin{tabular}{p{2.1cm}cccc}
\hline
Type & Prec & Recall & F1 \\
\hline\hline
\multicolumn{4}{c}{Experiments} \\
\hline 
\MWE[s]& 0.4697 & 0.4655 & 46.76\% \\
Supersenses & 0.5320 & 0.5138 & 52.27\% \\
Combined & 0.5194 & 0.5046 & 51.19\% \\
\hline
\multicolumn{4}{c}{Official} \\
\hline 
MWE& 0.6122 & 0.2807 &38.49\% \\
Supersenses& 0.5009 & 0.5326 & 51.62\% \\
Combined& 0.5114 & 0.4846 & 49.77\% \\
Macro& & & 49.94\% \\
\hline
\end{tabular}
\caption{Results}
\label{result}
\end{table}

\section {Results}

The system configuration which was used as our official run for the SemEval 2016 task was parameterized as follows: wide layer 1 in \MWE prediction perceptron = 1024 nodes; hash sum size = 16bit, calculated both for known and unknown word, using all characters in a word (alpha + non-alpha); a mean vector of word embeddings over the sentence was included as extra feature; distance features are supplied to the composed vector evaluator (as shown with the dotted line in \figref{arch}); a confidence level of $-0.15$ was applied to the \MWE perceptron when predicting a two-word \MWE prefix (but not when expanding it to the third and next words, encouraging an \MWE to start). The whole DiMSUM training set was used to train the system.

\tabref{result} displays the results, measured with the DiMSUM evaluation script.

\tabref{ablation} represents a brief ablation study with feature vector disabled. Word embeddings seem to be the critical source of supersense information, and they are also one of the important contributors to \MWE recall. The incremental $n$-word vector is critical for \MWE precision, and Distance features are of great importance both for precision and recall in \MWE detection, but especially for the recall. Word hash and, surprisingly, heuristic word features are of much less significance than other feature vectors.
\begin{table*}
\centering
\begin{tabular}{p{64pt}c@{\,\,\,}c@{\,\,\,}cc@{\,\,\,}c@{\,\,\,}cc@{\,\,\,}c@{\,\,\,}c}
\hline
Ablated &\multicolumn{3}{c}{\MWE[s]}&  \multicolumn{3}{c}{Supersenses} & \multicolumn{3}{c}{Combined}\\
features & Prec & Recall & F1 & Prec & Recall & F1 & Prec & Recall & F1\\
\hline\hline
$-$Recurrency&0.2125&0.4960&29.75\%&0.4638&0.3581&40.41\%&0.3591&0.3843&37.13\%\\
$-$Heuristic&0.4230&0.5821&48.99\%&0.5033&0.4797&49.12\%&0.4822&0.4991&49.05\%\\
$-$Distance&0.3246&0.2771&29.90\%&0.4494&0.4575&45.34\%&0.4281&0.4232&42.56\%\\
$-$Word hash&0.3640&0.6350&46.27\%&0.5133&0.4811&49.67\%&0.4674&0.5104&48.80\%\\
$-$word2vec &0.3204&0.5193&39.63\%&0.1898&0.1766&18.29\%&0.2293&0.2418&23.54\%\\
\hline
\end{tabular}
\caption{Feature ablation results}
\label{ablation}
\end{table*}

\section {Findings \& Conclusions}

The system captures supersenses rather well (when we consider the large number of senses, small size of the training data, and ambiguity in the sense assignments). However, it frequently admits harsh mispredictions that are probably caused by the lack of global sense coherence in \WE[s] \cite{qu2015big}. Also, it suffers from the lack of (other than \MWE-related) context information in cases of ambiguity. Improving context awareness may be the most obvious next step. Unsurprisingly, the most frequent supersenses have the best recall, up to 80\% for \sense{v.stative}. The mean recall value for senses is around 50--60\%, and there are senses (like \sense{n.weather}) that are never correctly predicted.  Exceptions are \sense{n.attribute}, \sense{n.location}, and \sense{v.change}, where recall is below 30\% despite them being reasonably frequent. Some of the most frequent mispredictions are the following: \sense{n.person} $\rightarrow$ \sense{n.group}; \sense{n.communication} $\rightarrow$ \sense{n.artefact}; \sense{n.attribut} $\rightarrow$ \sense{n.cognition}; \sense{n.act} $\rightarrow$ \sense{n.event}; and \sense{v.emotion} $\rightarrow$ \sense{v.cognition}.

MWE identification looks generally reasonable for all principal types of \MWE[s] (even, surprisingly, ones of an idiomatic nature), but the overall accuracy is low.
\subparagraph{Hysteresis bias pattern.}
The \MWE classifier tends to be biased toward \textit{not} joining
  two words into a \MWE \footnote{An example from the DiMSUM test data:
    \ex{English speaking advisor} produces an \MWE predicted to be \ex{English
    advisor} (not including \ex{speaking}}; at the same time, once an
  \MWE is predicted, it tends to be extended excessively to include a third and
  subsequent words, often selecting a non-relevant word with
  some gap.\footnote{As a very rough compensation of such an effect, one
    may use two confidence levels, one (negative) at $n=2$, another
    (positive) at $n\geq 3$.} This probably means that the proposed
  sampling schema is not balanced enough to work for small
  amounts of training data.
\subparagraph{Lookahead.}
The method obviously lacks the ability to model context when
  predicting \MWE[s]. In cases of expressions like \ex{John , Mary \&
  Company}, for example, our system will predict \ex{John
  Mary Company} with all the punctuation missing. Some bidirectional,
  attentional \cite{bahdanau2014neural,cohn2016incorporating}
  or lookahead-based approach is needed, as we may not have a
  reasonable isolated rule for whether \ex{John} should be joined to the comma. A similar situation may also occur in punctuation-less contexts.

\subparagraph{Parsing.}
The use of parsing results in markedly better precision and recall. Taking into account the strong correlation between
  parser-based features and \MWE boundaries, further investigation of the
  parser-based approach is warranted.

\subparagraph{Multiword-to-Sense Collision.}
A shared $n$-word incremental vector computation both for \MWE and sense
  learning imposes a collision. It may be observed that for better \MWE
  training, the distance features should be supplied to the computation
  input (as shown with the dotted line in \figref{arch}), but this will
  decreases the sense prediction score.

\subparagraph{Needs a deeper network.}
The training dynamics observed in the experiments show that the
  neural networks (especially the one used for \MWE prediction) need to be deeper
  (use more than two layers, as suggested in
  \newcite{sutskever2014sequence}).
\bibliography{papers}
\end{document}